%% file: main.tex
\documentclass{article}

\usepackage[preprint,nonatbib]{neurips_2020}

\usepackage[utf8]{inputenc} 
\usepackage[T1]{fontenc}    
\usepackage{hyperref}       
\usepackage{url}            
\usepackage{booktabs}       
\usepackage{amsfonts}       
\usepackage{nicefrac}       
\usepackage{microtype}      

\usepackage{algorithm}
\usepackage[noend]{algpseudocode}
\usepackage{amsthm,amssymb,amsmath}
\usepackage{graphicx}
\usepackage{subfig}
\usepackage{multicol}

\newtheorem{theorem}{Theorem}

\title{DREAM: Deep Regret Minimization with Advantage Baselines and Model-free Learning}

\author{
   \textbf{Eric Steinberger}\thanks{Facebook AI Research (FAIR)}
   \protect\phantom{\footnotesize 1}\thanks{ClimateScience (https://climate-science.com)}
   \and
   \textbf{Adam Lerer}\footnotemark[1]
   \and
   \textbf{Noam Brown}\footnotemark[1]
 }
 
\date{\vspace{-5ex}}
\begin{document}

\maketitle
\input{core/0_abstract.tex}

\input{core/1_introduction.tex}
\input{core/3_background.tex}
\input{core/2_related_work.tex}
\input{core/3a_cfr.tex}
\input{core/4_DREAM.tex}
\input{core/5_experiments.tex}
\input{core/6_results.tex}
\input{core/7_conclusions.tex}
\newpage
\input{core/8_Broader_Impact.tex}

\input{core/9_Ackn.tex}
\bibliographystyle{plain}
\bibliography{main}
\newpage
\appendix
\input{core/A_games.tex}
\input{core/C_DDQN_BR.tex}
\input{core/B_proofs.tex}

\input{core/E_NN.tex}

\end{document}

%% file: core/0_abstract.tex
\begin{abstract}
We introduce DREAM, a regret-based deep reinforcement learning algorithm that converges to an equilibrium in imperfect-information multi-agent settings. Our primary contribution is an effective algorithm that, in contrast to other regret-based deep learning algorithms, does not require access to a perfect simulator of the game in order to achieve good performance. We show that DREAM empirically achieves state-of-the-art performance among model-free algorithms in popular benchmark games, and is even competitive with algorithms that do use a perfect simulator.

\end{abstract}

%% file: core/1_introduction.tex
\section{Introduction}
\label{sec:intro}
We consider the challenge of computing a Nash equilibrium in imperfect-information two-player zero-sum games. This means finding a policy such that both agents play optimally against the other and no agent can improve by deviating to a different policy.

There has been an explosion of deep reinforcement learning algorithms that compute optimal policies in single-agent settings and perfect-information games~\cite{mnih2015human,double-dqn,dueling-dqn,schulman2017proximal}. However, these algorithms generally fail to compute an optimal policy in imperfect-information games.

More recently, deep reinforcement learning algorithms have been developed that either theoretically or empirically converge to a Nash equilibrium in two-player zero-sum imperfect-information games~\cite{heinrich2016deep,srinivasan2018actor,brown2019deep,steinberger2019single,li2020double,omidshafiei2019neural,perolat2020poincar}. Among these, neural forms of counterfactual regret minimization (CFR)~\cite{zinkevich2008regret,lanctot2009monte,tammelin2014solving,brown2019solving} have achieved the best performance~\cite{brown2019deep,steinberger2019single,li2020double}. This mirrors the success of tabular CFR, variants of which are state-of-the-art among tabular equilibrium-finding algorithms~\cite{brown2019solving} and which have been used in every major AI milestone in the domain of poker~\cite{bowling2015heads,moravvcik2017deepstack,brown2017superhuman,brown2019superhuman}.

However, in order to deploy tabular CFR in domains with very large state spaces, domain-specific abstraction techniques that bucket similar states together are needed~\cite{PotentialAwareAbstr,Habstr}. While these abstraction techniques have been successful in poker, they require extensive domain knowledge to construct and are not applicable to all games. This has motivated the development of CFR algorithms that use neural network function approximation to generalize with far less domain knowledge.

However, the existing forms of neural CFR only perform well with an exact simulator of the game, which allows them to sample and explore multiple actions at a decision point and thereby reduce variance. We show they perform poorly if only a single action is sampled, which is necessary in a model-free setting.

In this paper we introduce \emph{DREAM}, a neural form of CFR that samples only a single action at each decision point. In order to keep variance low despite sampling only a single action, DREAM uses a sample-based neural form of learned baselines, a fundamental technique that has previously been shown to be successful in tabular CFR~\cite{schmid2019variance,davis2019low}.

DREAM minimizes regret and converges to an $\epsilon$-Nash equilibrium in two-player zero-sum games with $\epsilon$ proportional to the neural modeling error.

We demonstrate empirically that DREAM achieves state-of-the-art performance among model-free self-play deep reinforcement learning algorithms in two popular benchmark poker variants: Leduc Hold'em and Flop Texas Hold'em.

%% file: core/3_background.tex
\section{Notation and Background}

Our notation is based on that of partially observable stochastic games~\cite{hansen2004dynamic}. We consider a game with $\mathcal{N} = \{1,2,...,N\}$ agents. We use $i$ to denote the agent index and use $-i$ to refer to all agents except agent~$i$.

A \textbf{world state} $w \in \mathcal{W}$ is the exact state of the world. $\mathcal{A} = \mathcal{A}_1 \times \mathcal{A}_2 \times ... \times \mathcal{A}_N$ is the space of joint actions. $\mathcal{A}_i(w)$ denotes the legal actions for agent~$i$ at $w$ and $a = (a_1, a_2, ..., a_N) \in \mathcal{A}$ denotes a joint action. After each agent chooses a legal action, a transition function $\mathcal{T}$ determines the next world state $w'$ drawn from the probability distribution $\mathcal{T}(w,a) \in \Delta \mathcal{W}$. In each world state $w$, agent~$i$ receives a reward $\mathcal{R}_i(w)$.
Upon transition from world state $w$ to $w'$ via joint action $a$, agent~$i$ receives an \textbf{observation}~$O_i$ from a function $\mathcal{O}_i(w,a,w')$.

A \textbf{history} (also called a \textbf{trajectory}) is a finite sequence of legal actions and world states, denoted $h = (w^0, a^0, w^1, a^1, ..., w^t)$. The reward $i$ obtained in the last world state of a history $h$ is $\mathcal{R}_i(h)$ and $i$'s legal actions are $\mathcal{A}_i(h)$. An \textbf{infostate} (also called an \textbf{action-observation history (AOH)}) for agent~$i$ is a sequence of an agent's observations and actions $s_i = (O_i^0, a_i^0, O_i^1, a_i^1, ..., O_i^t)$. The set of all infostates for agent~$i$ is $\mathcal{I}_i$. The unique infostate containing history $h$ for agent~$i$ is denoted $s_i(h)$. The set of histories that correspond to an infostate $s_i$ is denoted $\mathcal{H}(s_i)$. Note that all histories in an infostate are, by definition, indistinguishable to agent~$i$, allowing us to define $\mathcal{A}(s_i) = \mathcal{A}(h)$ and $\mathcal{R}(s_i) = \mathcal{R}(h)$ for all $h \in s_i$. An observation $O_i^t$ may and typically does include rewards achieved upon the state transitions as well as the opponents' actions chosen in joint actions leading to the observation. We denote $h' \sqsubset h$ to indicate h' is a history reached on the path to $h$.

An agent's \textbf{policy} $\pi_i$ is a function mapping from an infostate to a probability distribution over actions. A \textbf{policy profile} $\pi$ is a tuple $(\pi_1, \pi_2, ..., \pi_N)$. The policy of all agents other than $i$ is denoted $\pi_{-i}$. A policy for a history $h$ is denoted $\pi_i(h) = \pi_i(s_i(h))$ and $\pi(h) = (\pi_1(s_1(h)), \pi_2(s_2(h)),...,\pi_N(s_N(h)))$. We also define the transition function $\mathcal{T}(h, a_i, \pi_{-i})$ as a function drawing actions for $-i$ from $\pi_{-i}$ to form $a = a_i \cup a_{-i}$ and to then return the history $h'$ from $\mathcal{T}(w_{last}, a)$, where $w_{last}$ is the last world state in $h$.

The expected sum of future rewards (also called the \textbf{expected value (EV)}) for agent~$i$ in history $h$ when all agents play policy profile $\pi$ is denoted $v_i^{\pi}(h)$. The EV for an infostate $s_i$ is denoted $v_i^{\pi}(s_i)$ and the EV for the entire game is denoted $v_i(\pi)$. The EV for an action in a history and infostate is denoted $v_i^{\pi}(h,a_i)$ and $v_i^{\pi}(s_i,a_i)$, respectively. A \textbf{Nash equilibrium} is a policy profile such that no agent can achieve a higher EV by switching to a different policy. Formally, $\pi^*$ is a Nash equilibrium if for every agent~$i$, $v_i(\pi^*) = \max_{\pi_i}v_i(\pi_i, \pi^*_{-i})$. The \textbf{exploitability} of a policy profile $\pi$ is $e(\pi) = \sum_{i \in \mathcal{N}} \max_{\pi'_i}v_i(\pi'_i, \pi_{-i})$.

The \textbf{reach} $x^{\pi}(h)$ of a history $h$ is the product of all action probabilities under $\pi$ and all transition function probabilities leading to $h$. Formally, for $h^t = (w^0, a^0, w^1, a^1, ..., w^t)$, $x^{\pi}(h^t) = \Pi_{h^{n} \sqsubset h^t, i \in \mathcal{N}} \pi_i(h^n,a_i^n) \tau(w^n,a_i^n,w^{n+1})$.
The \textbf{agent reach} $x_i^\pi(h^t)$ of a history $h^t$ is the product of the probabilities for all agent~$i$ actions leading to $h^t$. Formally, $x_i^{\pi}(h^t) = \Pi_{h^n \sqsubset h^t}(h^n,a_i^n)$. We similarly define the agent reach $x_i^\pi(s_i)$ of infostate $s_i = (O_i^0,a_i^0,O_i^1,a_i^1,...,O_i^t)$ as $x_i^\pi(s_i) = \Pi_{t}(a_i^t)$.
The \textbf{external reach} $x_{-i}^{\pi}(s_i)$ of infostate $s_i$ is the probability that agent~$i$ would reach $s_i$ if they had always taken actions leading to $s_i$ with 100\% probability. Formally, $x_{-i}^{\pi}(s_i) = \sum_{h^t \in \mathcal{H}(s_i)} \Pi_{h^{n} \sqsubset h^t, i \in \mathcal{N} \setminus \{i\}} \pi_i(h^n,a_i^n)$. If $x_i^{\pi}(s_i) > 0$ then external reach is also $\sum_{h \in \mathcal{H}(s_i)} \frac{x^{\pi}(h)}{x_i^{\pi}(s_i)}$.

\label{sec:background}

%% file: core/2_related_work.tex
\section{Related Work}

\textit{Tabular} CFR methods are limited by the need to visit a given state to update the policy played in it~\cite{zinkevich2008regret,brown2019solving,tammelin2014solving,lanctot2009monte}. In large games that are infeasible to fully traverse, domain-specific abstraction schemes~\cite{PotentialAwareAbstr,Habstr} can shrink the game to a manageable size by clustering states into buckets. This \textit{abstracted} version of the game is solved during training and then mapped back onto the full game. These techniques work in many games~\cite{lisy2016equilibrium}, but they most often require expert knowledge to design well and are not applicable to all games.

To address these issues, researchers started to apply neural network function approximation to CFR. The first such experiment was \textit{DeepStack}~\cite{moravvcik2017deepstack}. While DeepStack used neural networks to predict a quantity called counterfactual value, it still relied on tabular solving for multiple stages of its training and evaluation process. The benefit of a parameterized policy is that it can make an educated guess for how to play in states it has never seen during training. Parameterized policies have led to performance breakthroughs in AI for perfect information games like Atari and Go~\cite{hessel2018rainbow,silver2017mastering}.

\textit{Neural Fictitious Self-Play (NFSP)}~\cite{heinrich2016deep} approximates extensive-form fictitious play using deep reinforcement learning from single trajectories. NFSP was the first deep reinforcement learning algorithm to learn a Nash Equilibrium in two-player imperfect information games from self-play. Since then, various policy gradient and actor critic methods have been shown to have similar convergence properties if tuned appropriately~\cite{srinivasan2018actor,lanctot2017unified}.

\textit{Deep CFR} is a fully parameterized variant of CFR that requires no tabular sampling~\cite{brown2019deep}. Deep CFR substantially outperformed NFSP (which in turn outperforms or closely matches competing algorithms~\cite{srinivasan2018actor}). \textit{Single Deep CFR} (SD-CFR)~\cite{steinberger2019single} removed the need for an average network in Deep CFR and thereby enabled better convergence and more time- and space-efficient training. However, both Deep CFR and Single Deep CFR rely on a perfect simulator of the game to explore multiple actions at each decision point.

\textit{Double Neural CFR} is another algorithm aiming to approximate CFR using deep neural networks~\cite{li2020double}. \textit{Advantage Regret Minimization (ARM)}~\cite{jin2017regret} is a regret-based policy gradient algorithm designed for single agent environments. Before neural networks were considered, \textit{Regression CFR (R-CFR)}~\cite{waugh2015solving} used regression trees to estimate regrets in CFR and CFR$^+$. However, R-CFR seems to be incompatible with sparse sampling techniques~\cite{srinivasan2018actor}, rendering it less scalable.

%% file: core/3a_cfr.tex
\section{Review of Counterfactual Regret Minimization (CFR)} \label{sec:review}
Counterfactual Regret Minimization (CFR)~\cite{zinkevich2008regret} is an iterative policy improvement algorithm that computes a new policy profile $\pi^t$ on each iteration~$t$. The average of these policies converges to a Nash equilibrium.

CFR can apply either \textbf{simultaneous} or \textbf{alternating} updates. If the former is chosen, CFR produces $\pi_i^t$ for all agents~$i$ on every iteration $t$. With alternating updates, CFR produces a policy only for one agent~$i$ on each iteration, with $i=$ $t$ $mod$ $2$ on iteration $t$. We describe the algorithm in this section in terms of simultaneous updates.

The \textbf{instantaneous regret} for action $a_i$ at infostate $s_i$ is \(r^t_i(s_i, a_i) = x_{-i}^{\pi^t}(s_i)\big(v_i^{\pi^t}(s_i, a_i) - v_i^{\pi^t}(s_i)\big)\). $r^t_i(s_i, a_i)$ is a measure of how much more agent~$i$ could have gained on average by always choosing $a_i$ in $s_i$, weighed by the external reach. The \textbf{average regret} on iteration $T$ is defined as \(R^T_i(s_i, a_i)=\frac{\sum^T_{t=1} r^t_i(s_i, a_i)}{T}\). CFR defines the policy for agent~$i$ on iteration $t+1$ based on their overall regret as
\begin{equation} \label{eq:vanCFRiterStrat}
    \pi^{t+1}_i(s_i, a_i) =
    \begin{cases}
        \frac{\max\{0,R^t_i(s_i, a_i)\}}{\sum_{a'_i\in \mathcal{A}_i(s_i)} \max\{0,R^t_i(s_i, a'_i)\}}  & \text{if} \sum_{a'_i \in \mathcal{A}_i(s_i)} \max\{0,R^t_i(s_i, a'_i)\} > 0 \\
        \frac{1}{|\mathcal{A}_i(s_i)|}                                     & \text{otherwise} \\
    \end{cases}
\end{equation}
The initial policy is set to uniform random. The \textbf{average policy} $\bar{\pi}_i^T$ is
\begin{equation} \label{eq:vanCFRavgStrat}
    \bar{\pi}^T_i(s_i, a_i) = \frac{\sum_{t=1}^{T} x_{i}^{\pi^t}(s_i)  \pi_{i}^{t}(s_i, a_i)}{\sum_{t=1}^{T} x_{i}^{\pi^t}(s_i)}
\end{equation}
On each iteration~$t$, CFR traverses the entire game tree and updates the regrets for every infostate in the game according to policy profile $\pi^t$. These regrets define a new policy $\pi^{t+1}$. The average policy over all iterations converges to a Nash equilibrium.

\subsection{Linear CFR}
Linear CFR~\cite{brown2019solving} is a variant of CFR that weighs the updates to the regret and average policy on iteration $t$ by $t$. Thus, under Linear CFR, \(R^T_{i}(s_i, a_i) = \sum^T_{t=1} (t r^t_i(s_i, a_i))\) and
\(\bar{\pi}^T_i(s_i, a_i) = \frac{\sum_{t=1}^{T} (t x_i^{\pi^t}(s_i)  \pi_i^{t}(s_i, a_i))} {\sum_{t=1}^{T} (t x_i^{\pi^t}(s_i))}\). This modification accelerates CFR by two orders of magnitude.

\subsection{Monte Carlo CFR (MC-CFR)}
\label{sec:MC-CFR}
Monte Carlo CFR (MC-CFR) is a framework that allows CFR to only update regrets on part of the tree for a single agent, called the \textbf{traverser}. Two variants relevant to our work are \textbf{External Sampling (ES)} and \textbf{Outcome Sampling (OS)}~\cite{lanctot2009monte}. In OS, regrets are updated for the traverser only along a single trajectory (i.e., a single action is sampled for each agent at each decision point, and a single outcome is sampled from $\mathcal{T}$). In ES, a single action is sampled for each non-traverser agent and a single outcome is sampled from $\mathcal{T}$, but every traverser action is explored. When agent~$i$ is the traverser in ES or OS, all other agents sample actions according to $\pi_{-i}$. In OS, we use a sampling policy~$\xi^t_i(s_i)$ profile rather than~$\pi^t$ to choose actions during play. $\xi^t_i(s_i)$ is defined separately for the traverser~$i$ and any other agent~$j$
\begin{equation} \label{eq:DREAMSamPoli}
\xi^{t}_i(s_i, a_i) = \epsilon\frac{1}{|\mathcal{A}_i(s_i)|} + (1-\epsilon)\pi^t_i(s_i, a_i)
\text{ }\text{ }\text{ }\text{ }\text{ }\text{ }\text{ }\text{ }\text{ }\text{ }\text{ }\text{ }\text{ }\text{ }\text{ }\text{ }\text{ }\text{ }\text{ }\text{ }\text{ }\text{ }\text{ }\text{ }\text{ }
\xi^{t}_{j}(s_j, a_j) = \pi^{t}_{j}(s_j, a_j)
\end{equation}
For a thorough and more general description of MC-CFR see~\cite{lanctot2009monte}.

\subsection{Variance-Reduced Outcome Sampling with Baselines (VR-MC-CFR)}
\label{sec:vrMC-CFR}
ES typically produces lower-variance estimates of expected values than OS. Variance-reduced MC-CFR (VR-MCCFR) is a method to reduce variance in MC-CFR and is particularly helpful for OS~\cite{schmid2019variance,davis2019low}. Given that OS ultimately reaches a terminal history $z$, the \textbf{baseline-adjusted sampled expected value} in history $h$ is
\begin{equation} \label{eq:VRMC-CFRva}
    \Tilde{v}_{i}^{\pi^t}(s_i(h),a_i|z) =
    \Tilde{v}_{i}^{\pi^t}(h,a_i|z) =
    \begin{cases}
        b_{i}(h, a_i) + \frac{\sum_{h'}\{p(h' | \mathcal{T}(h,a_i,\pi_{-i}^t))\Tilde{v}_{i}^{\pi^t}(h'|z)\} - b_i(h,a_i)}{\xi_i^t(s_i,a_i)} & \text{if } (h, a_i) \sqsubseteq z \\
        
        b_i(h, a_i) & \text{otherwise}
    \end{cases}
\end{equation}
\begin{equation} \label{eq:VRMC-CFRv}
    \Tilde{v}_{i}^{\pi^t}(s_i(h)|z) =
    \Tilde{v}_{i}^{\pi^t}(h|z) =
    \begin{cases}
        \mathcal{R}_i(h) & \text{if } h = z \\
        \sum_{a_i \in \mathcal{A}_i(h)} \pi_i^t(h,a_i) \Tilde{v}_{i}^{\pi^t}(h,a_i) & \text{otherwise} \\
    \end{cases}
\end{equation}
where $b$ is called the \textbf{history-action baseline}, a user-defined scalar function that ideally is closely correlated with $v$. Common choices in the tabular setting are a running average across visits to any given history, and domain-specific estimates of $v$~\cite{schmid2019variance,davis2019low}.

\subsection{Deep CFR and Single Deep CFR (SD-CFR)}
\label{sec:deepcfr}
Deep CFR~\cite{brown2019deep} approximates the behavior of tabular CFR from partial game tree traversals. For each CFR iteration~$t$, Deep CFR runs a constant $T$ external sampling traversals to collect samples of instantaneous regret and adds the data to a reservoir buffer $B^d_i$~\cite{reservoirSampling}, where agent~$i$ is the traverser. Deep CFR then trains a \textbf{value network}~$\hat{D}^t_i$ to predict the average regret (divided by external reach) for unseen infostate actions. $\hat{D}
^t_i$ is then used to approximate the policy $\pi^{t+1}_i$ for agent~$i$ like tabular CFR would produce given $R^t_i$. More data is collected by sampling based on this policy, and the process repeats until convergence.

Specifically, on each iteration $T$, the algorithm fits a value network $\hat{D}^T_i$ for one agent~$i$ to approximate \textbf{advantages} defined as \(D^T_i(s_i,a_i) = \frac{R^T_{i}(s_i, a_i)}{\sum_{t=1}^T (x_{-i}^{\pi^t}(s_i))}\). Deep CFR divides $R^T_i(s_i, a_i)$ by \(\sum_{t=1}^T (x_{-i}^{\pi^t}(s_i))\) to erase the difference in magnitude arising from highly varying realisation probabilities across infostates, which makes the data easier for the value network to approximate. This division does not change the policy output by Equation~\ref{eq:vanCFRiterStrat} since every action in an infostate is divided by the same value.

In cases where Deep CFR predicts non-positive advantage for all infostate actions, it chooses the action with the highest advantage rather than a uniform random policy as vanilla CFR would do.

Since it is the \emph{average} policy that converges to a Nash equilibrium in CFR, not the final strategy, Deep CFR trains a second neural network~$\hat{S}$ to approximate the average policy played over all iterations. Data for $\hat{S}_i$ is stored in a separate reservoir buffer $B^s_i$ and collected during the same traversals that data for $B^v_i$ is being collected on.

Single Deep CFR (SD-CFR) is a modification of Deep CFR that instead stores all value networks from each CFR iteration to disk and mimics the average policy exactly during play by sampling one of them and using its policy for the entire game. This is mathematically equivalent to sampling actions from the average policy. SD-CFR eliminates the approximation error in Deep CFR resulting from training a network to predict the average policy, at the minor cost of using extra disk space to store the models from each CFR iteration (or a random sample of CFR iterations).

%% file: core/4_DREAM.tex
\section{Description of DREAM}
In this section, we will describe DREAM, (D)eep (RE)gret minimization with (A)dvantage Baselines and (M)odel-free learning. The primary contribution of this paper is to effectively combine a deep neural network approximation of outcome sampling CFR with learned baselines.

Like Single Deep CFR, DREAM is trained using alternating CFR iterations~\cite{burch2019revisiting}. On each iteration $t$ at each encountered infostate~$s_i$, we obtain estimated advantages $\hat{D}^t_i(s_i,a_i|\theta_i^t)$ for the actions~$a_i \in \mathcal{A}_i(s_i)$ by inputting $s_i$ into the advantage network parameterized by $\theta_i^t$ . From these estimated advantages, we obtain a policy $\pi_i^t(s_i)$ via regret matching as shown in Equation~\ref{eq:vanCFRiterStrat}, except using $\hat{D}
^t_i$ in place of $R_i^t$, and picking the action with the highest $\hat{D}^t_i$ probability 1 when all advantages are negative.

\subsection{Incorporating outcome sampling with exploration}
On each iteration $t$, DREAM collects advantages from $T$ trajectories of the game for agent~$i$, where $i=$ $t$ $\mod$ $N$ following the outcome sampling (OS) MC-CFR scheme~\cite{lanctot2009monte} with the sampling policy profile introduced in \ref{sec:vrMC-CFR}.
where $\epsilon > 0$ is the exploration parameter. To compensate for the shifted sampling distribution due to $\epsilon$, we weigh data samples stored in $B^d_i$ by \(\frac{1}{x^{\xi^t}_i(s_i)}\) during neural training. Storing advantage models from each iteration and sampling one randomly to determine a policy used during play, as is done in SD-CFR, results in an algorithm we call Outcome Sampling SD-CFR (OS-SD-CFR). We show in section~\ref{sec:results} that OS-SD-CFR performs poorly.

\subsection{Variance reduction using a learned Q-baseline}
We hypothesised that the higher variance of advantage estimates in OS traversals is to blame for OS-SD-CFR's bad performance. DREAM aims to reduce this source of variance by implementing a neural form of history baselines~\cite{davis2019low}, the tabular form of which is described in section~\ref{sec:vrMC-CFR}.

DREAM uses a baseline network \(\hat{Q}^t_i(s^*(h), a_i | \phi^t_i)\) parameterized by $\phi^{t}_i$ as a baseline, where $s^*$ is the set of infostates for all players at $h$, $s^*(h)=(s_0(h),...,s_N(h))$.\footnote{We do not define $\hat{Q}$ to take $h$ directly as input since $h$ is not available in a model-free setting. However, in the games we investigate, $h$ and $s^*(h)$ are equivalent.} $\hat{Q}^t_i$ is fine-tuned from $\hat{Q}^{t-1}_i$ when $t > 0$ and initialized with random weights on $t=0$. It is trained using expected SARSA~\cite{van2009theoretical} on a circular buffer $B_i^q$. Formally, it updates $\phi^t_i$ to minimize its mean squared error (MSE) to the target value sample
\(
    \mathcal{R}_i(h) + \sum_{a_i' \in \mathcal{A}_i} \pi^{t}(s_i(h'),a_i')\hat{Q}^t_i(s^*(h'), a_i')
\),
where $h'$ is drawn from $\mathcal{T}(h,a_i,\pi^t_{-i})$.

VR-MCCFR computes history-action baselines both for player actions and `chance actions' from the environment. The latter requires access to a perfect simulator of the environment's transition function. Thus, DREAM only uses baselines for the player actions. Given that OS ultimately reaches a terminal history $z$, the \textbf{baseline-adjusted sampled expected value} at history $h$ where action $a_i'$ is played leading to $h'$, is
\begin{equation} \label{eq:DREAMua}
    \Tilde{v}_{i,DREAM}^{\pi^t}(h,a_i|z) = 
    \begin{cases}
        \hat{Q}^t_{i}(s^*(h), a_i) - \frac{\Tilde{v}_{i,DREAM}^{\pi^t}(h'|z) - \hat{Q}^t_i(s^*(h),a_i)}{\xi^t(s_i,a_i)} & \text{if } a_i = a_i' \\
        
        \hat{Q}^t_i(s^*(h), a_i) & \text{otherwise}
    \end{cases}
\end{equation}
\begin{equation} \label{eq:VRMC-CFRu}
    \Tilde{v}_{i,DREAM}^{\pi^t}(h|z) =
    \begin{cases}
        \mathcal{R}_i(h) & \text{if } h = z \\
        \sum_{a_i \in \mathcal{A}_i(h)} \pi_i^t(h,a_i) \Tilde{v}_{i,DREAM}^{\pi^t}(h,a_i|z) & \text{otherwise} \\
    \end{cases}
\end{equation}
and $\Tilde{v}_{i,DREAM}^{\pi^t}(s_i(h),a_i|z) = \Tilde{v}_{i,DREAM}^{\pi^t}(h,a_i|z)$ and $\Tilde{v}_{i,DREAM}^{\pi^t}(s_i(h)|z) = \Tilde{v}_{i,DREAM}^{\pi^t}(h|z)$.

We define the \textbf{sampled immediate advantage} as
\begin{equation} \label{eq:DREAMVRadv}
    \Tilde{d}_{i, DREAM}^{t}(s_i,a_i)
    = \Tilde{v}_{i, DREAM}^{\pi^t}(s_i,a_i) - \Tilde{v}_{i, DREAM}^{\pi^t}(s_i)
\end{equation}
Note that the expectation of $\Tilde{d}_{i, DREAM}^{t}(s_i,a_i)$ is $\frac{r_i^t(s_i,a_i)}{x_{-i}^{\pi^t}(s_i,a_i)}$. Because we defined the sampling policy for $-i$ as \(\xi^{t}_{-i}(s_i, a_i) = \pi^{t}_{-i}(s_i, a_i)\), the expectation of samples added to the advantage buffer $B^d_i$ on iteration $t$ in history $h$ will be proportional to $r_i^t(s_i(h),a_i)$. Furthermore, because from agent~$i$'s perspective histories are sampled based on their information, the expectation for data added to $B^d_i$ in infostate $s_i$ is proportional to $r_i^t(s_i,a_i)$. Thus, training $\hat{D}^t_i$ on $B^d_i$ means it approximates a quantity proportional to the average regret as defined in section \ref{sec:review}. To implement Linear CFR~\cite{brown2019solving} in DREAM, we weigh the neural training loss of $\Tilde{d}^t_i$ in $B^d_i$ by $t$.

\subsection{The average policy}
In order to play according to the average policy, we follow the approach used in SD-CFR~\cite{steinberger2019single}, described in Section~\ref{sec:deepcfr}, in which the advantage model from each iteration, or a random subset thereof, is stored on disk in a set $B^M_i$ and one is sampled randomly at test time. That model is then used to determine the policy for the whole game. When implementing linear CFR, the probability of sampling model~$t$ is proportional to $t$. When needed, the explicit average policy probabilities can be computed in $\mathcal{O}(|B_i^M|)$ time by evaluating and averaging strategies derived from all networks in $B_i^M$.

Deep CFR's method of computing the average policy by training a separate neural network on action probability vectors collected during training~\cite{brown2019deep} is also compatible, but to correct for sampling bias, the neural training loss of samples from iteration $t$ is weighted by \(\frac{1}{x_{-i}^{\xi}}(h)\).

\subsection{Convergence}
The convergence of MC-CFR~\cite{lanctot2013monte} using function approximation to estimate infoset advantages has been previously studied in~\cite{brown2019deep}. They show that minimizing mean-squared error of predicted advantages on historical samples (as we do in DREAM) is equivalent to tabular MC-CFR. Furthermore, if a model only \textit{approximately} minimizes the MSE, its convergence bound incurs an additional constant average regret term proportional to the maximum gap $\epsilon_{\mathcal{L}}$ between the model's MSE error on the samples and the error of the exact minimizer (\cite{brown2019deep}, Theorem 1).

MC-CFR convergence results hold for any unbiased estimator of state-action values, and thus permit the use of arbitrary state-action baselines, although the rate of convergence is related to the variance of the baseline-adjusted estimator (i.e. the quality of the baseline). More details can be found in \cite{schmid2019variance}. We derive a bound on the cumulative regret for DREAM in Appendix \ref{app:proof}.


%% file: core/5_experiments.tex
\section{Experimental Setup}

We explore a sweep of hyperparameters for DREAM, and compare its exploitability over the number of game states observed to Deep CFR~\cite{brown2019deep}, Single Deep CFR (SD-CFR)~\cite{steinberger2019single}, and Neural Fictitious Self-Play (NFSP)~\cite{heinrich2016deep}. We chose Leduc poker~\cite{Leduc} and Flop hold'em poker~\cite{brown2019deep}, two popular poker games, as benchmark domains. A description of these can be found in the appendix. We use Linear CFR~\cite{brown2019solving} for DREAM and all variants of SD-CFR we evaluate. For all experiments, we plot the mean and standard deviation across three independent runs. Exploitability is measured in milli big blinds per game (mbb/g).

In FHP we measure an approximate lower bound on exploitability computed by training a DDQN agent~\cite{dueling-dqn,double-dqn} against DREAM and running a number of hands between DREAM and the final iteration of DDQN and for evaluation. Details are in appendix~\ref{app:rlbr}.

\subsection{Neural Networks}
Our work does not focus on exploring new neural architectures. Instead, to enable easy reproducibility, we use the neural architecture demonstrated to be successful with Deep CFR and NFSP~\cite{brown2019deep} for all networks in all algorithms and games. Details and notes on the required minimal modifications to input and output layers that were required for algorithmic reasons can be found in appendix \ref{app:nnarch}.

On each iteration, $\hat{Q}^t_i$ is trained for 1000 minibatches of 512 samples using the Adam optimizer~\cite{kingma2014adam} with a learning rate of 0.001 and gradient norm clipping of 1 in both games. $B^q_i$ has a capacity of 200,000. For comparibility, DREAM's value networks use the same training parameters as those SD-CFR~\cite{brown2019deep,steinberger2019single}, unless otherwise stated. Both SD-CFR and DREAM use advantage buffers $B^d_i$ with capacities of 40 million samples in FHP and 2 million in Leduc. Value networks for 3,000 batches of size 2,048 and 10,000 batches of size in Leduc and FHP, respectively. For DREAM with Deep CFR's average network we allow an additional buffer $B^s_i$ of 2 million samples per player. The average network is trained for 4,000 batches of 2,048 samples.

The number of traversals sampled per iteration for SD-CFR and DREAM was chosen such that both algorithms visit roughly the same number of states per iteration. In Leduc, we chose a successful value for DREAM (900) and then divide by 2.6 (the empirical average ratio between states visited by ES and OS in Leduc) to get 346 for SD-CFR. We compare this to the initially proposed setting of 1,500, which performs worse~\cite{steinberger2019single}. In FHP, we use 10,000 for SD-CFR as proposed by~\cite{brown2019deep} and 50,000 for DREAM. Again, the factor of 5 difference cancels out the ratio of states seen by OS compared to ES, determined empirically in preliminary runs. However, in FHP we stared from the parameters for SD-CFR and adjusted DREAM accordingly.

Apart from the neural network architecture, NFSP uses the same hyperparamters in Leduc as presented the original paper~\cite{heinrich2016deep}. Because the authors did not use FHP as a benchmark, we use the hyperparameters they applied in Limit hold'em poker, a similar game.

%% file: core/6_results.tex
\section{Experimental Results}
\label{sec:results}
\begin{figure}[!htb]
  \centering
  \captionsetup[subfigure]{labelformat=empty}
  \subfloat{\includegraphics[width=0.45\textwidth]{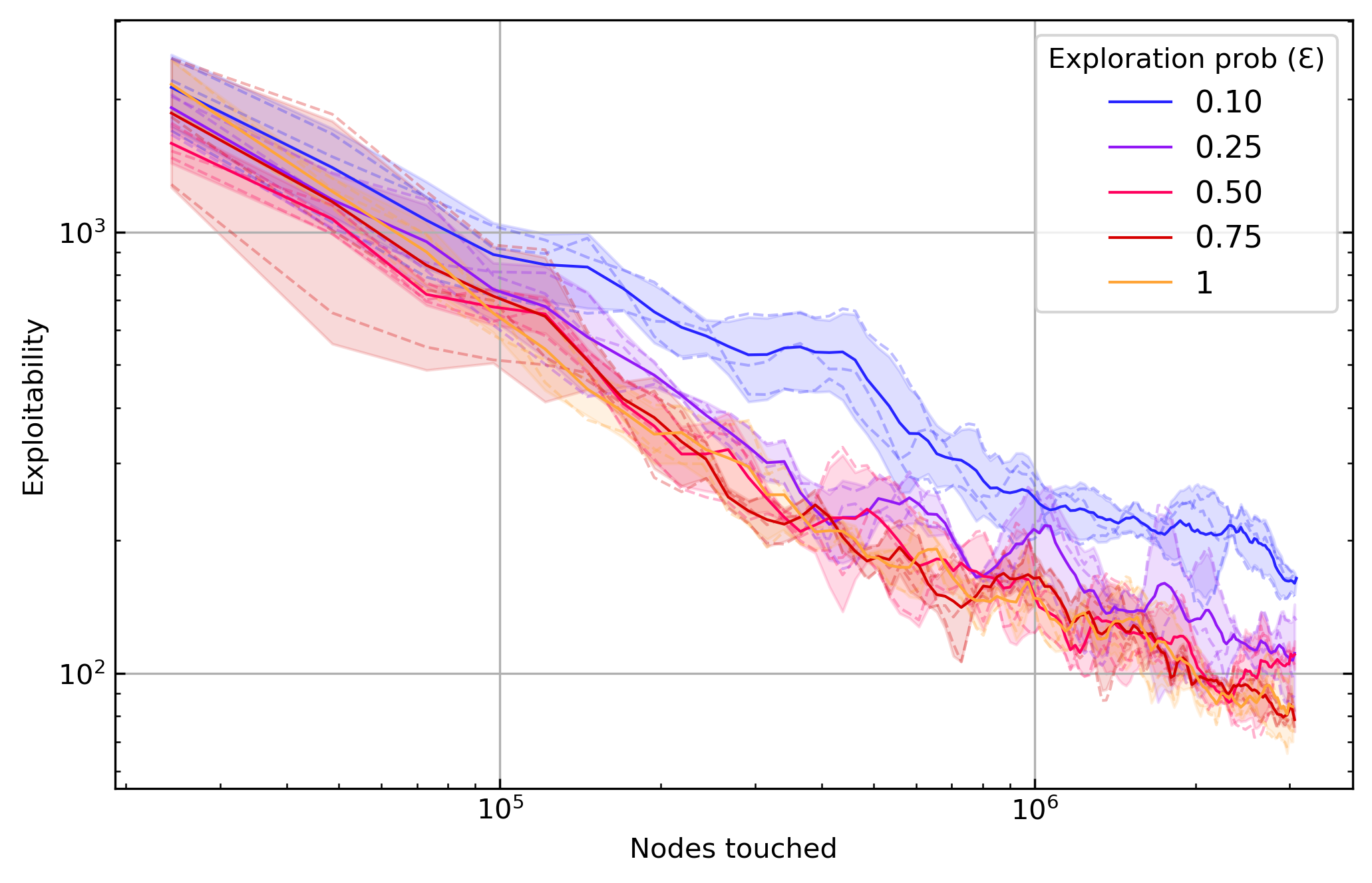}}
  \hfill
  \subfloat{\includegraphics[width=0.45\textwidth]{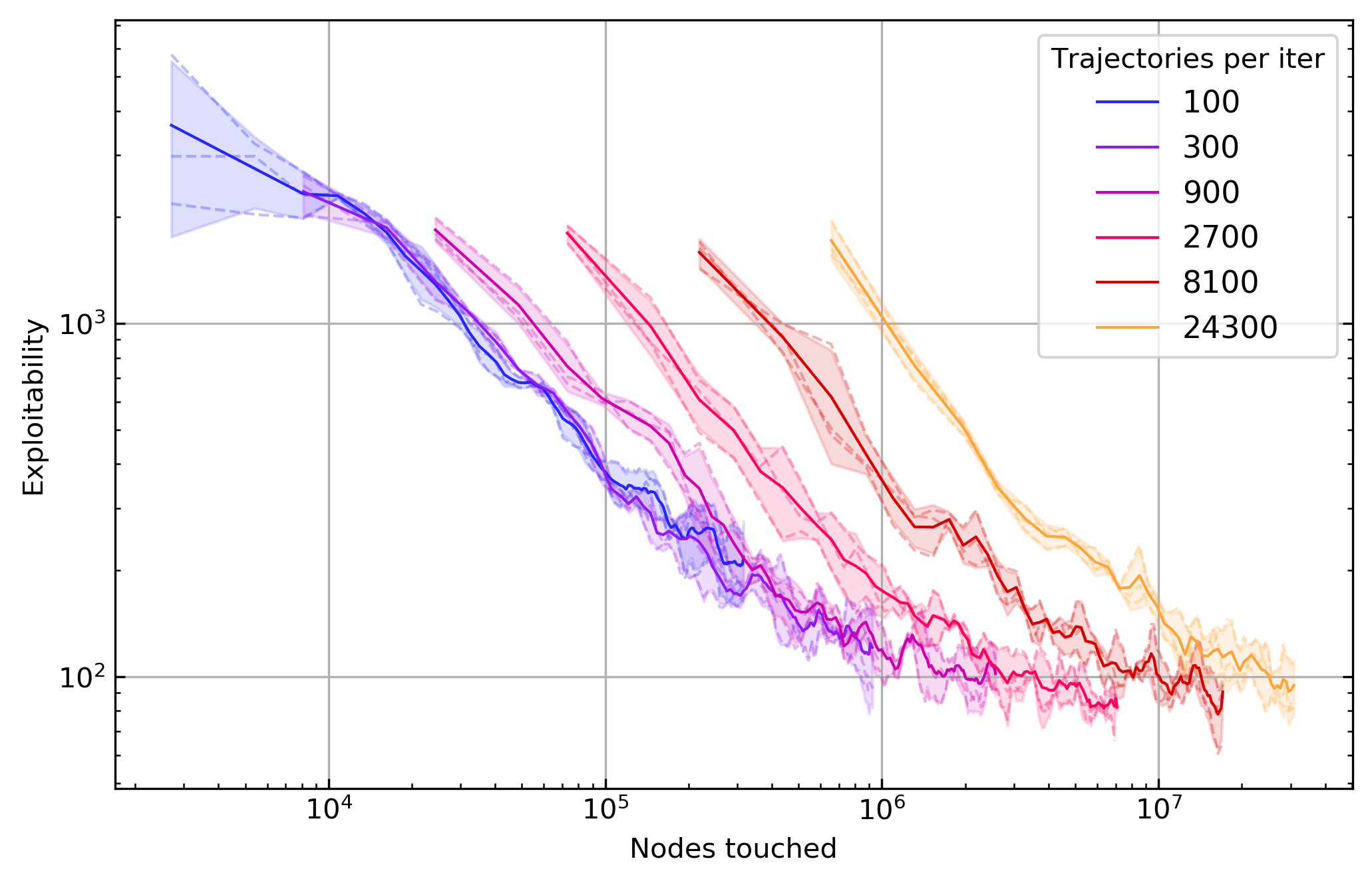}}
  \hfill
  \subfloat{\includegraphics[width=0.45\textwidth]{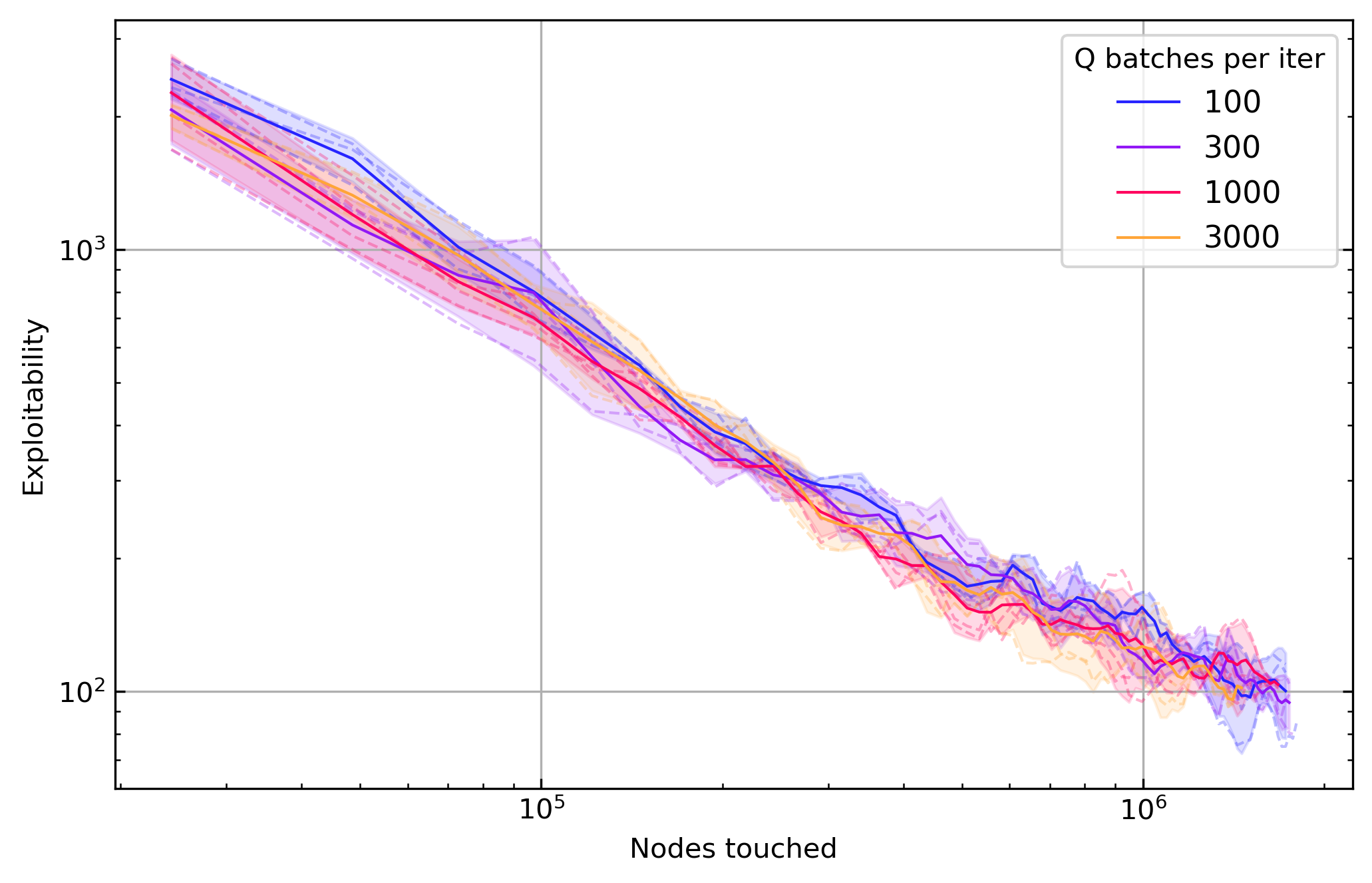}}
  \hfill
  \subfloat{\includegraphics[width=0.45\textwidth]{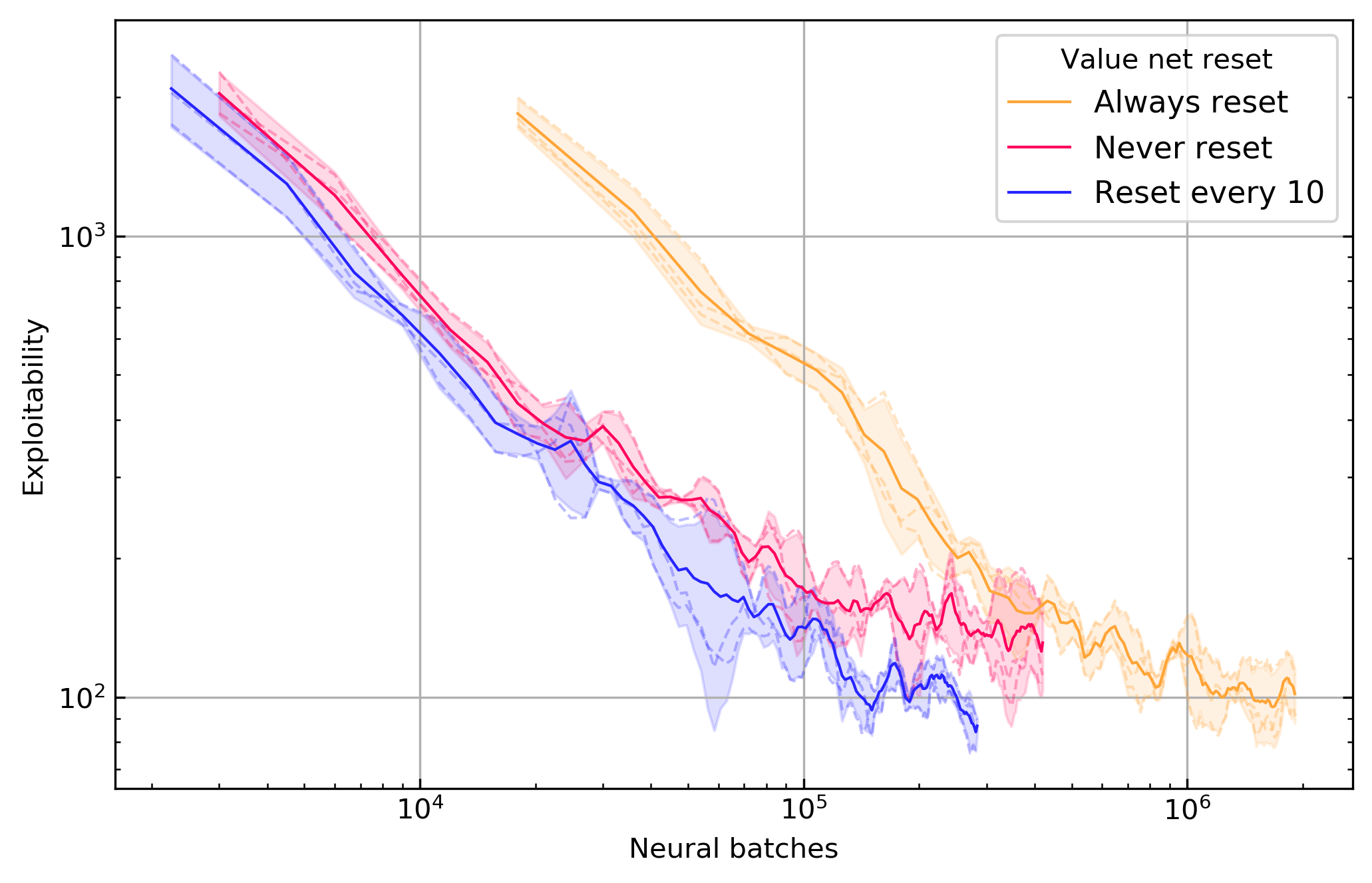}}  
  \caption{\small{Ablation studies in Leduc.
  \textbf{Top-left}: Comparing $\epsilon$ for DREAM.
  \textbf{Top-right}: Various numbers of trajectories sampled per DREAM iteration. \textbf{Bottom-left}: Various numbers of minibatch updates of DREAM's $\hat{Q}$ per iteration. \textbf{Bottom-right}: Modes of resetting the parameters of $\hat{D}$ in DREAM.}}
  \label{fig:LeducExperiments}
\end{figure}

\begin{figure}[!htb]
  \centering
  \captionsetup[subfigure]{labelformat=empty}
  \subfloat{\includegraphics[width=0.45\textwidth]{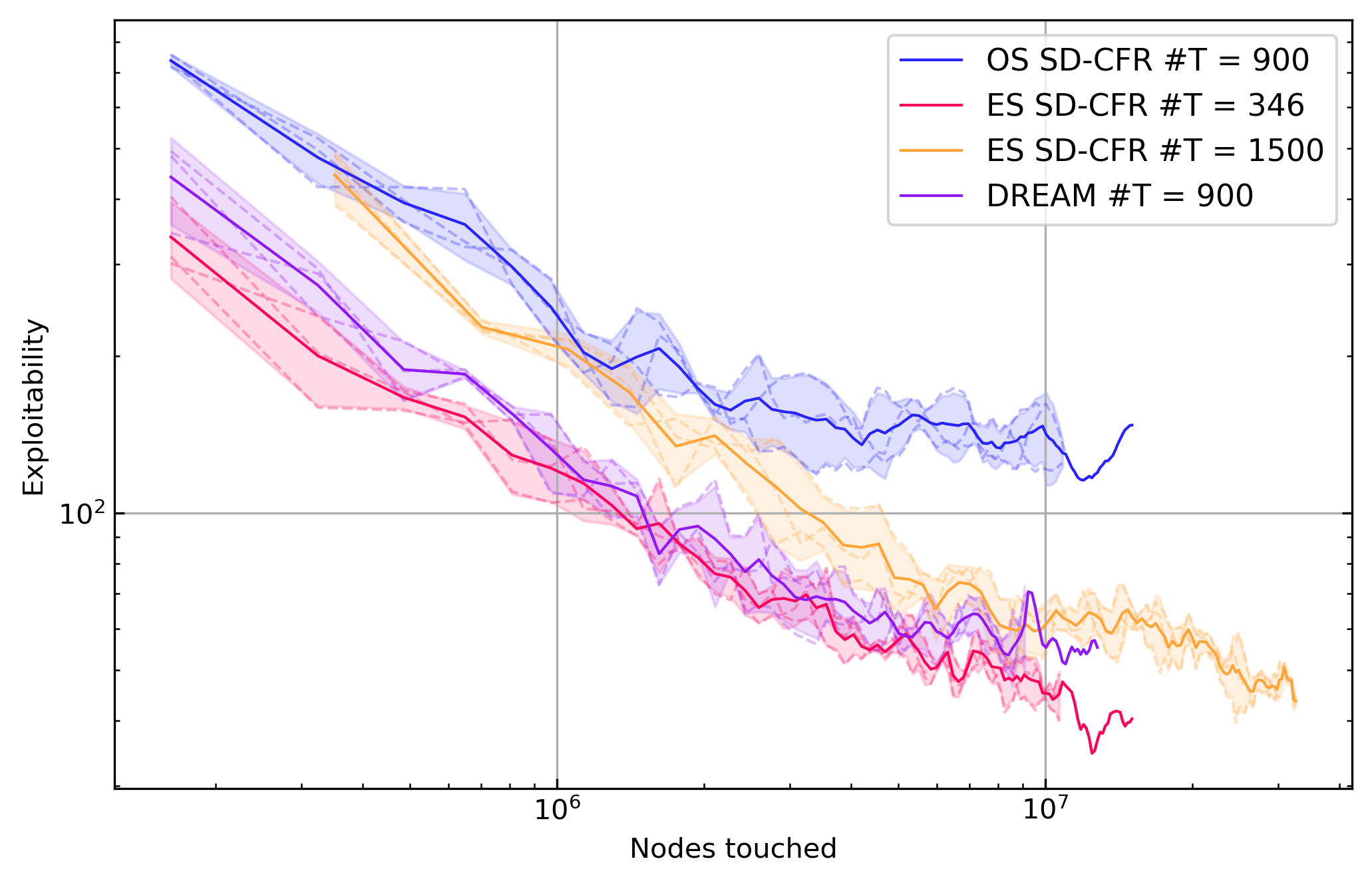}}
  \hfill
  \subfloat{\includegraphics[width=0.45\textwidth]{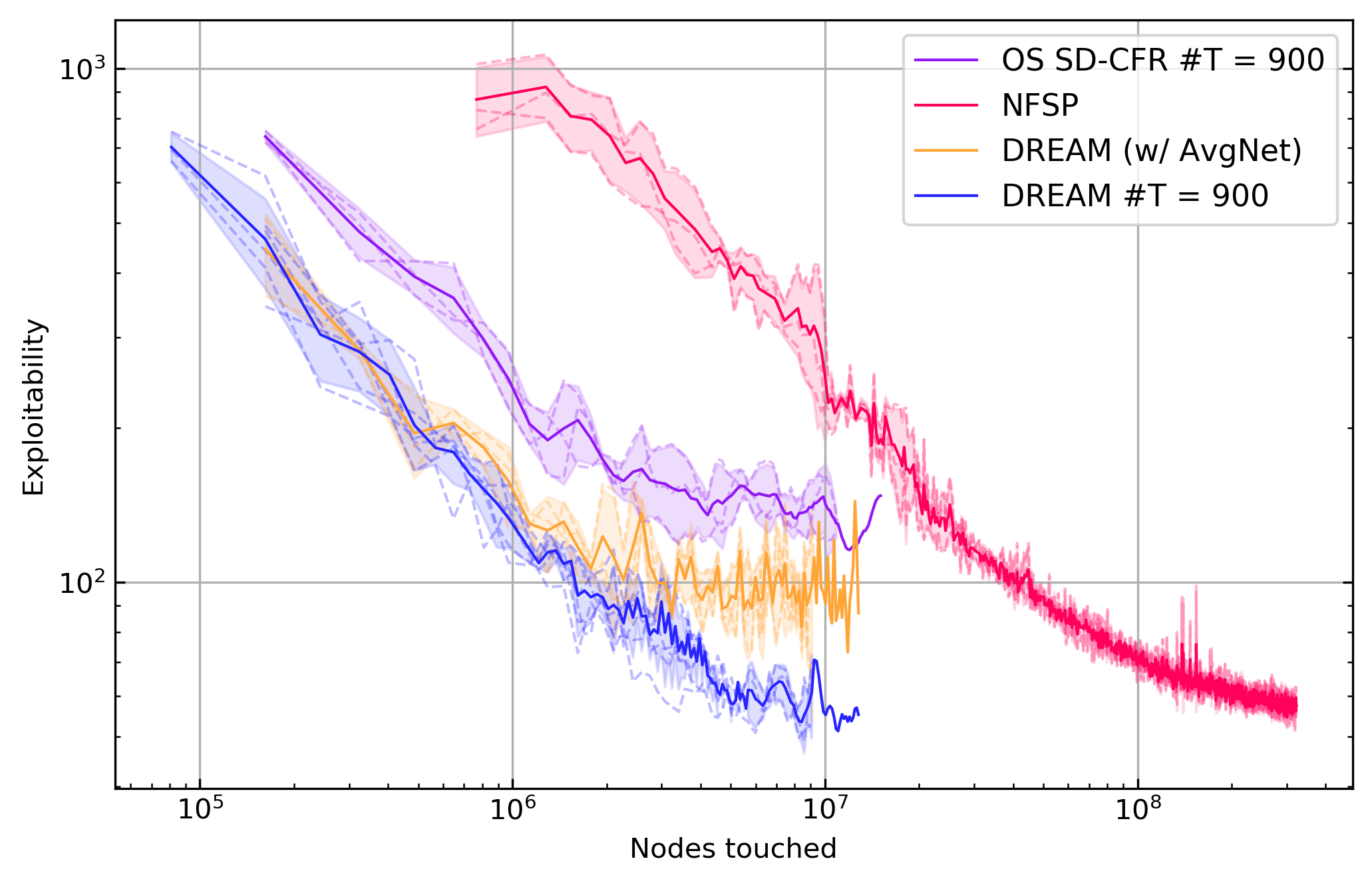}}
  \hfill
  \subfloat{\includegraphics[width=0.45\textwidth]{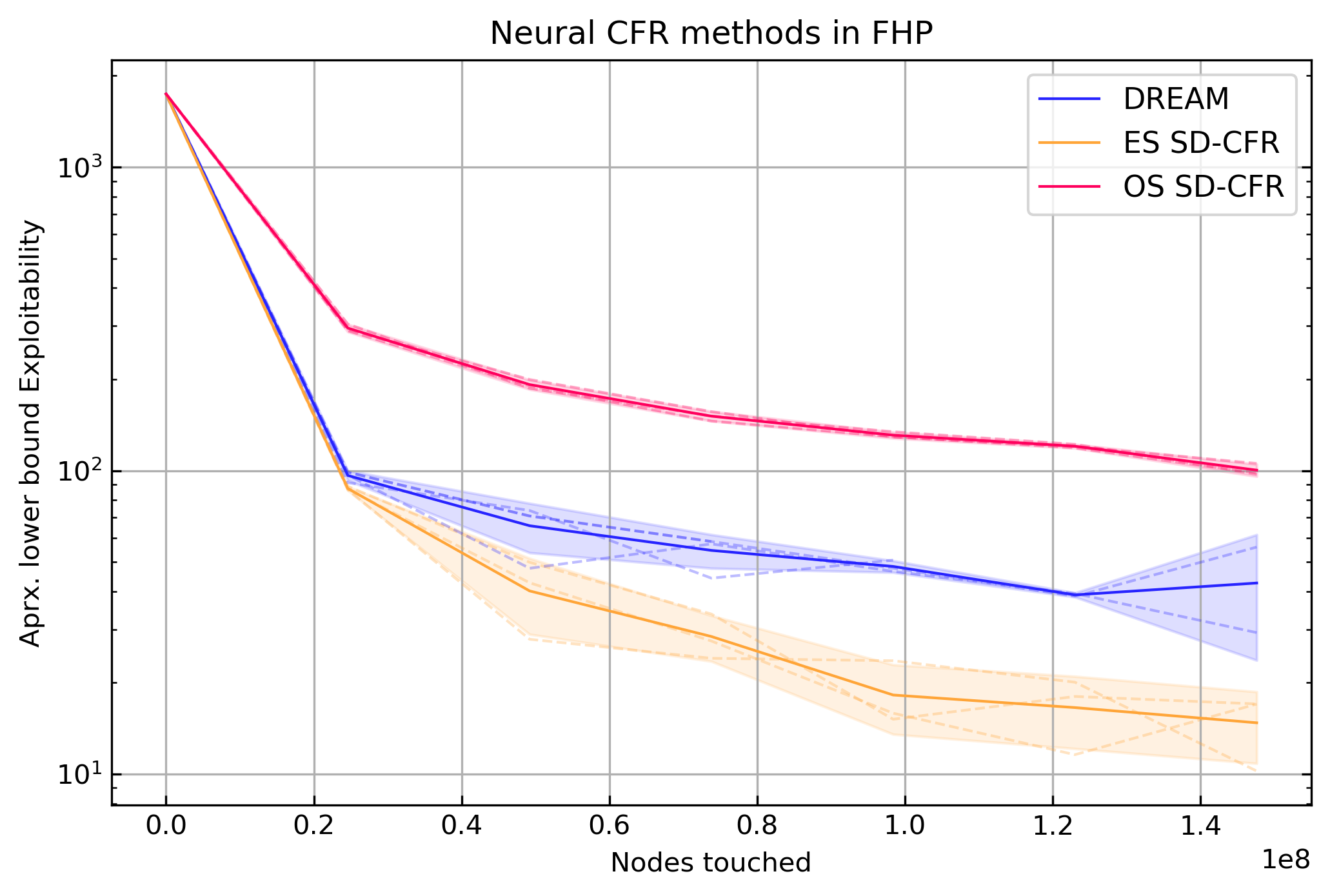}}
  \hfill
  \subfloat{\includegraphics[width=0.45\textwidth]{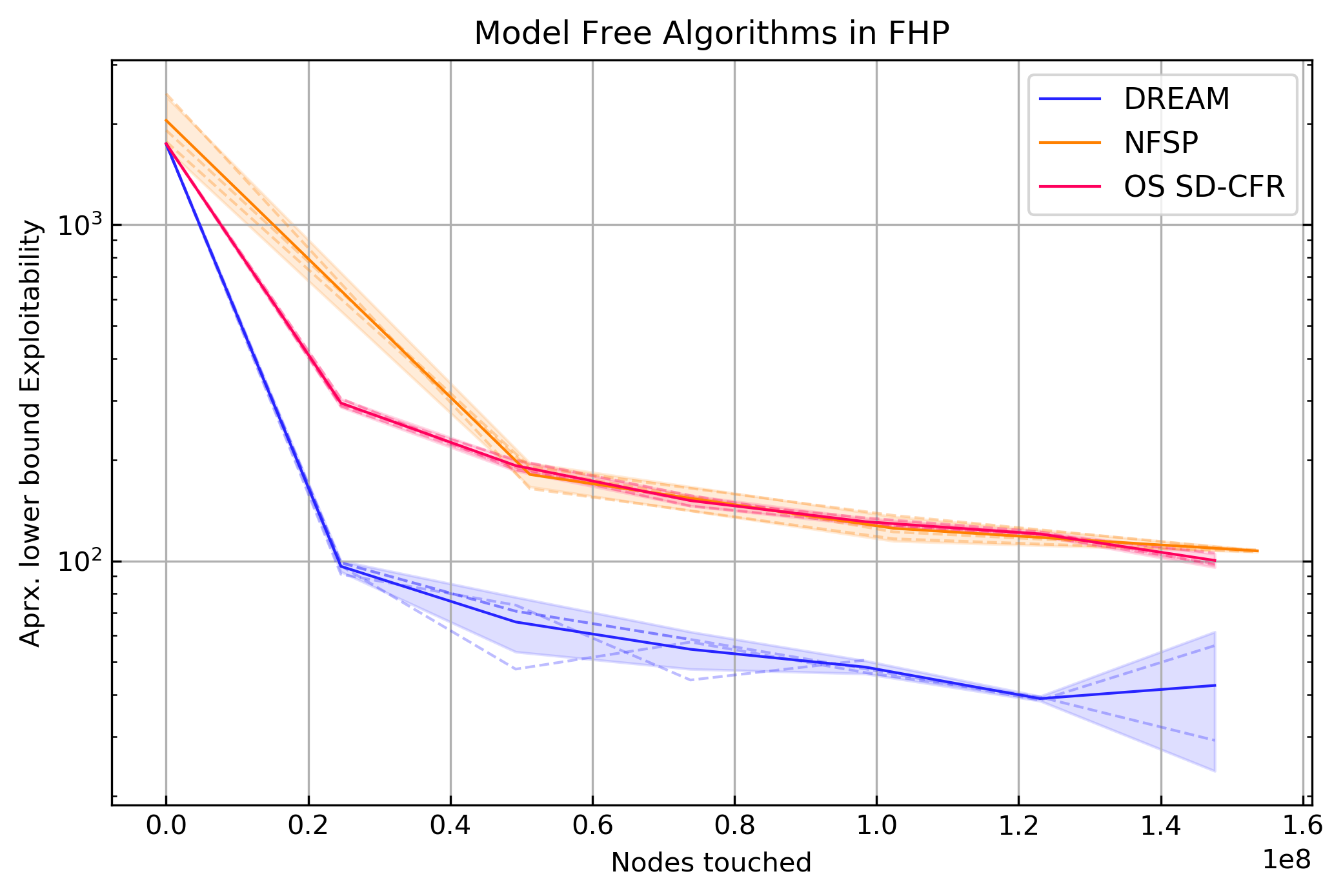}}
  \caption{\small{Exploitability comparison of DREAM with other algorithms in Leduc poker (top) and Flop Hold'em Poker (bottom). DREAM performs competitively with non-model-free Neural CFR methods (left), and outperforms all other model-free neural methods (right). The number of traversals for SD-CFR variants were chosen such that the number of nodes traversed per iteration is approximately equal to that of DREAM.}}
  \label{fig:LeducAndFHPAlgos}
\end{figure}

We first investigate settings and hyperparameters of DREAM (see figure~\ref{fig:LeducExperiments}) and then compare DREAM to other algorithms (see figure~\ref{fig:LeducAndFHPAlgos}). We find that DREAM significantly outperforms other model-free algorithms in Leduc and FHP. DREAM's performance matches that of SD-CFR in Leduc, despite not requiring a perfect simulator of the game.

We found that few minibatch updates per iteration are needed for $\hat{Q}$ in DREAM. Thus, the overhead relative to OS-SD-CFR that the baseline network adds is not significant. This may be because training targets for $\hat{Q}$ change only slightly on each iteration and $\hat{Q}^t$ is started from $\hat{Q}^{t-1}$'s weights. Moreover, small modelling errors may be insignificant compared to the remaining game variance.

Deep CFR~\cite{brown2019deep} performs better when value networks $\hat{D}^t$ are trained from scratch on each iteration $t$. We investigate three settings: 1) \textbf{Always reset}: train from scratch for 3,000 minibatches on every iteration, 2) \textbf{Never reset}: train from scratch for 3,000 minibatches on $t=0$ but train 500 minibatches starting from $\hat{D}^{t-1}_i$ if $t>0$, 3) \textbf{Reset every 10 iterations}: train from scratch for 3,000 minibatches if $t$ $mod$ $10 = 0$ else finetune for just 500 minibatches starting from $\hat{D}^{t-1}_i$.

%% file: core/7_conclusions.tex
\section{Conclusions}
\vspace{-0.1in}
We have introduced DREAM, a model-free self-play deep reinforcement learning algorithm. DREAM converges to a Nash equilibrium in two-player zero-sum imperfect-information games. In practise, DREAM achieves state-of-the-art performance among model-free RL algorithms in imperfect information games, beating the existing baseline NFSP~\cite{heinrich2016deep} by two orders of magnitude with respect to sample complexity in Leduc Hold'em and reaching a far lower exploitability than NFSP after multiple days of training in Flop Hold'em Poker.

%% file: core/8_Broader_Impact.tex
\section{Broader Impact} \label{sec:broaderImpact}
DREAM seeks to find equilibria in imperfect-information games without relying on an explicit model of the interaction. In contrast to most other works in deep reinforcement learning, our algorithms are aiming to find \emph{equilibrium} policies. This is particularly relevant for iterative decision-making processes - both competitive~\cite{brown2019solving} and cooperative~\cite{bard2019hanabi}.

Beneficial future applications of DREAM after further research and development could include political and market management decision support, negotiation and agreement in self-driving car fleets, and other decision making processes. Equilibrium computation is an essential part of robust multi-agent decision making.

Potential adverse applications most prominently include providing easy access to tools that aid cheating in online imperfect-information games with our open-sourced implementation of DREAM. For poker, the most prominent imperfect-information game and the one where cheating would have the most significant consequences, successful domain-specific algorithms already exist~\cite{brown2019superhuman}, so this publication should not add significant additional risk in that domain.

%% file: core/9_Ackn.tex
\section{Acknowledgements}
We thank Yannis Wells for providing feedback on drafts of this paper and interesting discussions.

%% file: core/A_games.tex
\section{Rules of Flop Hold'em Poker}
Flop Hold'em Poker (FHP) is a two-player zero-sum game. The positions of the two players
alternate after each hand. When acting, players can choose to either fold, call, or raise. When a player folds, the money in the pot is awarded to the other player. When a player calls, the player places money in the pot to equal to the opponent’s contribution. Raising means adding more chips to the pot than the opponent’s contribution.

A round ends when a player calls (and both players have acted). There can be at most three raises in each betting round. Raises are always in increments of \$100.
At the start of a hand, players are dealt two private cards from a standard 52-card deck. P1 must contribute \$50 to the pot and P2 must contribute \$100. A betting round then occurs starting with P1. After the round ends, three community cards are revealed face up. Another round of betting starts, but P2 acts first this time. At the end of this betting round, if no player has folded, the player with the strongest five-card poker hand wins, where the player's two private cards and the three community cards are taken as their five-card hand. In the case of a tie, the pot is split evenly between both players.

\section{Rules of Leduc Poker}
Leduc Poker differs from Flop Hold'em poker in that the deck only six cards, made from two suits $\{a, b\}$ and three ranks $\{J, Q, K\}$. Like FHP, the game consists of two rounds. At the start of the game, each player contributes \$50, called the ante, to the pot and is handed one private card. There can be at most two raises per round, where the bet-size is fixed at \$100 in the first round, and \$200 in the second. When the game transitions from the first to the second betting round, one public card is revealed. If no player folded, the strongest hand wins. If thee rank of a player's private card matches that of the public card, they win. If not, $K > Q > J$ determines the strongest hand. If both players' private cards have the same rank, the pot is split between them.

%% file: core/C_DDQN_BR.tex
\section{Approximate lower bound on the Best Response using RL} \label{app:rlbr}

When agent $i$'s policy is frozen, computing a best response policy for $-i$ can be seen as a partially observable single-agent problem. We leverage this by running a deep RL algorithm to make best response computations feasible in bigger games. While almost any deep RL algorithm could be used, we chose DDQN~\cite{double-dqn,dueling-dqn}. We train DDQN for 50,000 iterations with a batch size of 2048 and 512 steps per iteration. The circular buffer can hold 400,000 infostates. The target network is updated every 300 iterations. DDQN's exploration parameter decays exponentially from 0.3 to 0.02.

After training, we run 1 million hands between the agent and the learned approximate BR to obtain a \textbf{lower bound on the actual BR} with minimal uncertainty (95\% confidence interval of 5mBB/G in our experiments). This computation is run on 6 CPUs and takes 12 hours per datapoint.

\begin{figure}[!htb]
  \centering
  \includegraphics[width=0.45\textwidth]{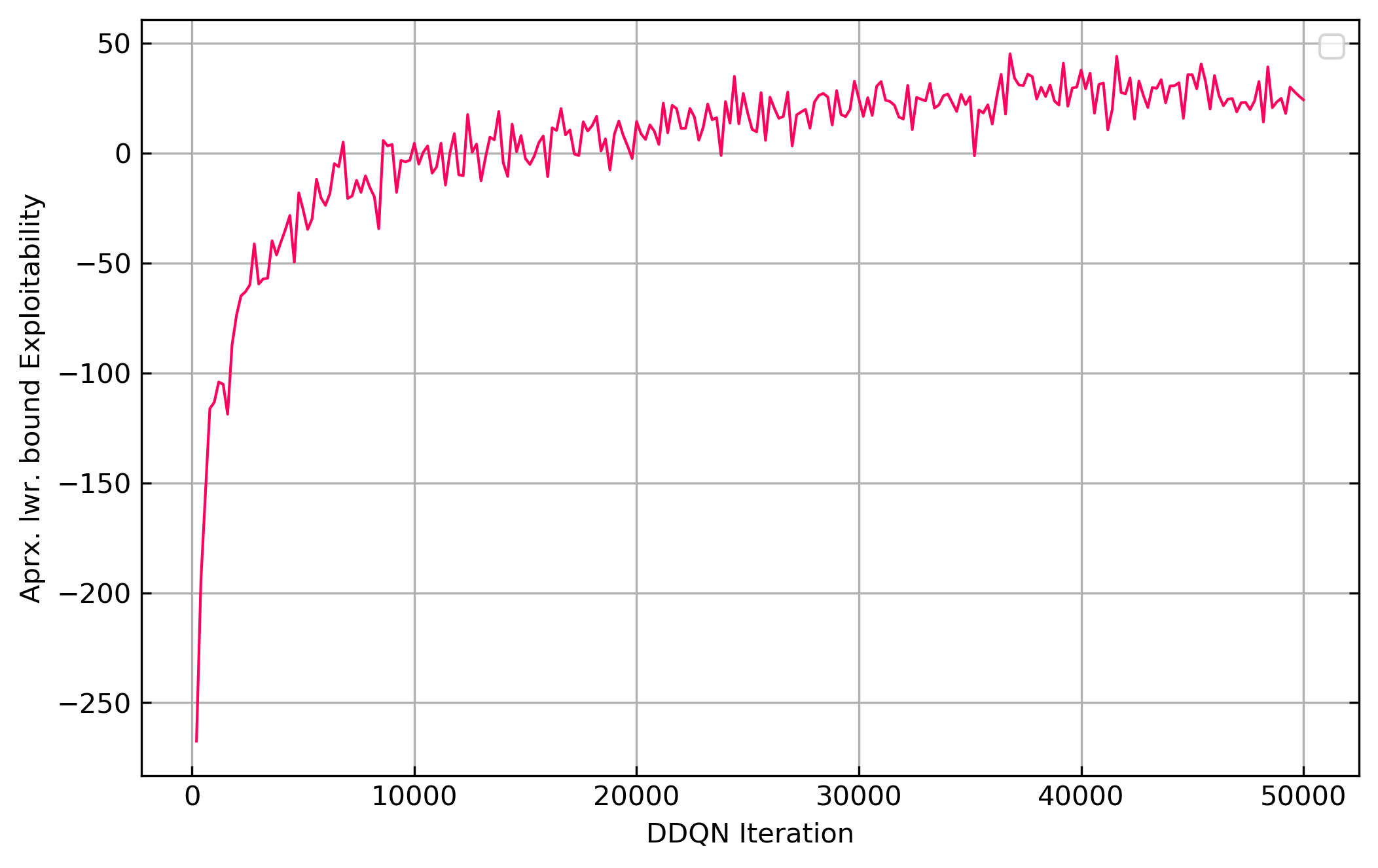}
  \caption{Convergence of DDQN as the approximate best response against DREAM on iteration 180 in Flop Hold'em Poker.}
  \label{fig:RLBR}
\end{figure}

Figure \ref{fig:RLBR} shows the winnings during training, average across both agents. This sample was taken on DREAM iteration 180 (out of around 500) and is representative of the general trend. Note that this includes exploration and is thus not representative of the approximate BR computed in the end, where no exploration is done.

We also found by running multiple instances of DDQN that the variance in this evaluation is fairly low, with all DDQN replicas arriving at similar exploitabilities.

%% file: core/B_proofs.tex
\section{Proofs of Theorems}
\label{app:proof}

\subsection{Mapping notation principles}
Our notation is more closely related to that of the reinforcement learning community than the conventional notation in the CFR literature. Most importnatly, what's usually referred to as $A$ in CFR ~\cite{lanctot2009monte} is $\mathcal{A}_i$ in our notation. The transition function $\mathcal{T}$ used in our work is usually referred to as $chance$ in the conventional notation. Moreover, the conventional notation assumes only one player acts in each node. We generalise this by considering opponent actions part of the transition function from the perspective of agent~$i$.

\subsection{Bound on DREAM's regret}
This section is written in the notation typical for work on extensive form games. Due to the relation between the notation used in our paper and that used in prior work on extensive form games, the references in this proof remain theoretically sound~\cite{kovavrik2019rethinking}.

\begin{theorem}
Suppose a series of policies $\pi^t$ are chosen via a CFR learning rule with outcome sampling, using the $\epsilon$-regret-matching sampling policy $\xi_i^t$ for the traversing agent described in \ref{eq:DREAMSamPoli}. Let $d_i$ be the maximum number of sequential decision points for agent~$i$ (i.e. the game depth for $i$). For any $p \in [0, 1)$, the average regret $R_i^T$ for agent~$i$ over this series of policies is bounded by

\begin{equation}
R_i^T \leq \frac{\Delta}{\sqrt{T}} 
\left( |\mathcal{I}_i| \sqrt{|\mathcal{A}_i|} + 2
    \left(\frac{|\mathcal{A}_i|}{\epsilon}\right)^{d_i} 
    \sqrt{2 \log{\frac{1}{p}}} 
\right)
\end{equation}
\end{theorem}

with probability $1-p$.

\begin{proof}

Let $R_i^T$ and $\tilde{R}_i^T$ be the average regret under MC-CFR and CFR, respectively.

We start from the original CFR bound in \cite{zinkevich2008regret}, Theorem 4: If agent~$i$ selects actions according to CFR then $\tilde{R}_i^T \leq \Delta |\mathcal{I}_i| \frac{\sqrt{|\mathcal{A}_i|}}{\sqrt{T}}$.

Next, we apply the bound from \cite{farina2020stochastic}, Proposition 1:  If agent~$i$ selects actions according to MC-CFR then

\begin{equation}
P\left[ R^T(u) \leq \tilde{R}^T(u) + (M + \tilde{M}) \sqrt{\frac{2}{T} \log{\frac{1}{p}}} \right] \geq 1 - p
\end{equation}

where $M$ and $\tilde{M}$ are bounds on the difference between the minimum and maximum loss in the full and sampled game, respectively. Trivially, $M \leq \Delta$. According to \cite{farina2020stochastic} Lemma 2, if agent~$i$ plays some sampling policy $\xi_i^t$ and the opponent agents their true policy $\pi_i^t$, then $\tilde{M} \leq \max_{z\in \mathcal{Z}}{\frac{\Delta}{\xi_i^t[z]}}$, where $\xi_i^t[z]$ is the probability that $\xi_i^t$ plays to terminal node $z$ assuming the opponent and chance do.

DREAM uses an $\epsilon$-regret matching sampling scheme for agent~$i$; therefore, we can easily bound $\xi_i^t[z] \leq \left( \frac{\epsilon}{|\mathcal{A}_i|} \right)^{d_i}$, where $d_i^{max}$ is the maximum number of sequential decision points for agent~$i$ on any path of the game.

Putting this all together, we find that for any $p \in [0, 1)$, 

\begin{equation}
    R_i^T \leq \Delta |\mathcal{I}_i| \frac{\sqrt{|\mathcal{A}_i|}}{\sqrt{T}} + \left(\Delta + \Delta \left( \frac{|\mathcal{A}_i|}{\epsilon} \right)^{d_i} \right) \sqrt{\frac{2}{T} \log{\frac{1}{p}}}
\end{equation}

with probability $1-p$.

\end{proof}

Importantly, this theorem does not imply that the actions chosen by the MC-CFR learning procedure are themselves no-regret; indeed, since DREAM uses an off-policy sampling scheme its regret during "training" may be arbitrarily high. The theorem instead implies that the historical average of the CFR policies $\pi^t$, which is the policy returned by the DREAM algorithm, converges to a Nash equilibrium at this rate in two-player zero-sum games.

%% file: core/E_NN.tex
\section{Neural Network Architecture} \label{app:nnarch}
\begin{figure}[h]
  \centering
  \includegraphics[width=0.8\textwidth]{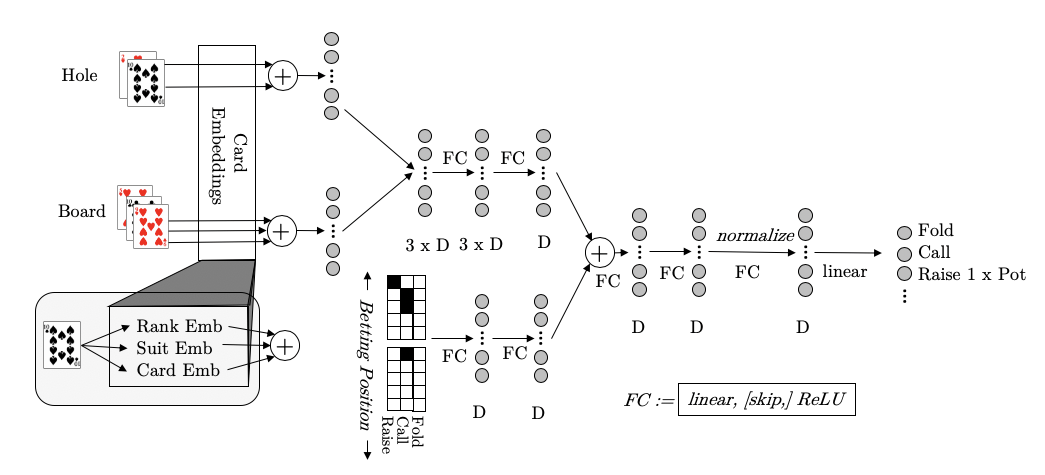}
  \caption{Neural network architecture as in~\cite{brown2019deep}. We set D=64 in all experiments.}
  \label{fig:NNBody}
\end{figure}

We use the neural architecture demonstrated to be successful with Deep CFR and NFSP~\cite{brown2019deep} (see figure \ref{fig:NNBody}). Slight differences between the networks are that we apply a softmax function to the last layer of NFSP's and Deep CFR's average networks like the original papers do. Moreover, DREAM's $Q$ network requires the input of both player's private cards and thus has a slightly adjusted input layer.

%% file: main.bbl
\begin{thebibliography}{10}

\bibitem{bard2019hanabi}
Nolan Bard, Jakob~N Foerster, Sarath Chandar, Neil Burch, Marc Lanctot,
  H~Francis Song, Emilio Parisotto, Vincent Dumoulin, Subhodeep Moitra, Edward
  Hughes, et~al.
\newblock The {H}anabi challenge: A new frontier for {A}{I} research.
\newblock {\em arXiv preprint arXiv:1902.00506}, 2019.

\bibitem{bowling2015heads}
Michael Bowling, Neil Burch, Michael Johanson, and Oskari Tammelin.
\newblock Heads-up limit hold’em poker is solved.
\newblock {\em Science}, 347(6218):145--149, 2015.

\bibitem{Habstr}
Noam Brown, Sam Ganzfried, and Tuomas Sandholm.
\newblock Hierarchical abstraction, distributed equilibrium computation, and
  post-processing, with application to a champion no-limit texas hold'em agent.
\newblock In {\em Proceedings of the 2015 International Conference on
  Autonomous Agents and Multiagent Systems}, pages 7--15. International
  Foundation for Autonomous Agents and Multiagent Systems, 2015.

\bibitem{brown2019deep}
Noam Brown, Adam Lerer, Sam Gross, and Tuomas Sandholm.
\newblock Deep counterfactual regret minimization.
\newblock In {\em International Conference on Machine Learning}, pages
  793--802, 2019.

\bibitem{brown2017superhuman}
Noam Brown and Tuomas Sandholm.
\newblock Superhuman {A}{I} for heads-up no-limit poker: Libratus beats top
  professionals.
\newblock {\em Science}, page eaao1733, 2017.

\bibitem{brown2019solving}
Noam Brown and Tuomas Sandholm.
\newblock Solving imperfect-information games via discounted regret
  minimization.
\newblock In {\em Proceedings of the AAAI Conference on Artificial
  Intelligence}, volume~33, pages 1829--1836, 2019.

\bibitem{brown2019superhuman}
Noam Brown and Tuomas Sandholm.
\newblock Superhuman {A}{I} for multiplayer poker.
\newblock {\em Science}, page eaay2400, 2019.

\bibitem{burch2019revisiting}
Neil Burch, Matej Moravcik, and Martin Schmid.
\newblock Revisiting cfr+ and alternating updates.
\newblock {\em Journal of Artificial Intelligence Research}, 64:429--443, 2019.

\bibitem{davis2019low}
Trevor Davis, Martin Schmid, and Michael Bowling.
\newblock Low-variance and zero-variance baselines for extensive-form games.
\newblock {\em arXiv preprint arXiv:1907.09633}, 2019.

\bibitem{farina2020stochastic}
Gabriele Farina, Christian Kroer, and Tuomas Sandholm.
\newblock Stochastic regret minimization in extensive-form games, 2020.

\bibitem{PotentialAwareAbstr}
Sam Ganzfried and Tuomas Sandholm.
\newblock Potential-aware imperfect-recall abstraction with earth mover's
  distance in imperfect-information games.
\newblock In {\em AAAI}, pages 682--690, 2014.

\bibitem{hansen2004dynamic}
Eric~A Hansen, Daniel~S Bernstein, and Shlomo Zilberstein.
\newblock Dynamic programming for partially observable stochastic games.
\newblock In {\em AAAI}, volume~4, pages 709--715, 2004.

\bibitem{heinrich2016deep}
Johannes Heinrich and David Silver.
\newblock Deep reinforcement learning from self-play in imperfect-information
  games.
\newblock {\em arXiv preprint arXiv:1603.01121}, 2016.

\bibitem{hessel2018rainbow}
Matteo Hessel, Joseph Modayil, Hado Van~Hasselt, Tom Schaul, Georg Ostrovski,
  Will Dabney, Dan Horgan, Bilal Piot, Mohammad Azar, and David Silver.
\newblock Rainbow: Combining improvements in deep reinforcement learning.
\newblock In {\em Thirty-Second AAAI Conference on Artificial Intelligence},
  2018.

\bibitem{jin2017regret}
Peter Jin, Kurt Keutzer, and Sergey Levine.
\newblock Regret minimization for partially observable deep reinforcement
  learning.
\newblock {\em arXiv preprint arXiv:1710.11424}, 2017.

\bibitem{kingma2014adam}
Diederik~P Kingma and Jimmy Ba.
\newblock Adam: A method for stochastic optimization.
\newblock {\em arXiv preprint arXiv:1412.6980}, 2014.

\bibitem{kovavrik2019rethinking}
Vojt{\v{e}}ch Kova{\v{r}}{\'\i}k, Martin Schmid, Neil Burch, Michael Bowling,
  and Viliam Lis{\`y}.
\newblock Rethinking formal models of partially observable multiagent decision
  making.
\newblock {\em arXiv preprint arXiv:1906.11110}, 2019.

\bibitem{lanctot2013monte}
Marc Lanctot.
\newblock {\em Monte Carlo sampling and regret minimization for equilibrium
  computation and decision-making in large extensive form games}.
\newblock PhD thesis, University of Alberta, 2013.

\bibitem{lanctot2009monte}
Marc Lanctot, Kevin Waugh, Martin Zinkevich, and Michael Bowling.
\newblock Monte carlo sampling for regret minimization in extensive games.
\newblock In {\em Advances in neural information processing systems}, pages
  1078--1086, 2009.

\bibitem{lanctot2017unified}
Marc Lanctot, Vinicius Zambaldi, Audrunas Gruslys, Angeliki Lazaridou, Karl
  Tuyls, Julien P{\'e}rolat, David Silver, and Thore Graepel.
\newblock A unified game-theoretic approach to multiagent reinforcement
  learning.
\newblock In {\em Advances in Neural Information Processing Systems}, pages
  4190--4203, 2017.

\bibitem{li2020double}
Hui Li, Kailiang Hu, Shaohua Zhang, Yuan Qi, and Le~Song.
\newblock Double neural counterfactual regret minimization.
\newblock In {\em International Conference on Learning Representations}, 2020.

\bibitem{lisy2016equilibrium}
Viliam Lisy and Michael Bowling.
\newblock Equilibrium approximation quality of current no-limit poker bots.
\newblock {\em arXiv preprint arXiv:1612.07547}, 2016.

\bibitem{mnih2015human}
Volodymyr Mnih, Koray Kavukcuoglu, David Silver, Andrei~A Rusu, Joel Veness,
  Marc~G Bellemare, Alex Graves, Martin Riedmiller, Andreas~K Fidjeland, Georg
  Ostrovski, et~al.
\newblock Human-level control through deep reinforcement learning.
\newblock {\em Nature}, 518(7540):529--533, 2015.

\bibitem{moravvcik2017deepstack}
Matej Morav{\v{c}}{\'\i}k, Martin Schmid, Neil Burch, Viliam Lis{\`y}, Dustin
  Morrill, Nolan Bard, Trevor Davis, Kevin Waugh, Michael Johanson, and Michael
  Bowling.
\newblock Deepstack: Expert-level artificial intelligence in heads-up no-limit
  poker.
\newblock {\em Science}, 356(6337):508--513, 2017.

\bibitem{omidshafiei2019neural}
Shayegan Omidshafiei, Daniel Hennes, Dustin Morrill, Remi Munos, Julien
  Perolat, Marc Lanctot, Audrunas Gruslys, Jean-Baptiste Lespiau, and Karl
  Tuyls.
\newblock Neural replicator dynamics.
\newblock {\em arXiv preprint arXiv:1906.00190}, 2019.

\bibitem{perolat2020poincar}
Julien Perolat, Remi Munos, Jean-Baptiste Lespiau, Shayegan Omidshafiei, Mark
  Rowland, Pedro Ortega, Neil Burch, Thomas Anthony, David Balduzzi, Bart
  De~Vylder, et~al.
\newblock From poincar$\backslash$'e recurrence to convergence in imperfect
  information games: Finding equilibrium via regularization.
\newblock {\em arXiv preprint arXiv:2002.08456}, 2020.

\bibitem{schmid2019variance}
Martin Schmid, Neil Burch, Marc Lanctot, Matej Moravcik, Rudolf Kadlec, and
  Michael Bowling.
\newblock Variance reduction in monte carlo counterfactual regret minimization
  (vr-mccfr) for extensive form games using baselines.
\newblock In {\em Proceedings of the AAAI Conference on Artificial
  Intelligence}, volume~33, pages 2157--2164, 2019.

\bibitem{schulman2017proximal}
John Schulman, Filip Wolski, Prafulla Dhariwal, Alec Radford, and Oleg Klimov.
\newblock Proximal policy optimization algorithms.
\newblock {\em arXiv preprint arXiv:1707.06347}, 2017.

\bibitem{silver2017mastering}
David Silver, Julian Schrittwieser, Karen Simonyan, Ioannis Antonoglou, Aja
  Huang, Arthur Guez, Thomas Hubert, Lucas Baker, Matthew Lai, Adrian Bolton,
  et~al.
\newblock Mastering the game of go without human knowledge.
\newblock {\em Nature}, 550(7676):354, 2017.

\bibitem{Leduc}
Finnegan Southey, Michael~P Bowling, Bryce Larson, Carmelo Piccione, Neil
  Burch, Darse Billings, and Chris Rayner.
\newblock Bayes' bluff: Opponent modelling in poker.
\newblock {\em Proceedings of the 21st Conference in Uncertainty in Artificial
  Intelligence}, pages 550–--558, 2005.

\bibitem{srinivasan2018actor}
Sriram Srinivasan, Marc Lanctot, Vinicius Zambaldi, Julien P{\'e}rolat, Karl
  Tuyls, R{\'e}mi Munos, and Michael Bowling.
\newblock Actor-critic policy optimization in partially observable multiagent
  environments.
\newblock In {\em Advances in neural information processing systems}, pages
  3422--3435, 2018.

\bibitem{steinberger2019single}
Eric Steinberger.
\newblock Single deep counterfactual regret minimization.
\newblock {\em arXiv preprint arXiv:1901.07621}, 2019.

\bibitem{tammelin2014solving}
Oskari Tammelin.
\newblock Solving large imperfect information games using cfr+.
\newblock {\em arXiv preprint arXiv:1407.5042}, 2014.

\bibitem{double-dqn}
Hado van Hasselt, Arthur Guez, and David Silver.
\newblock Deep reinforcement learning with double q-learning.
\newblock {\em CoRR}, abs/1509.06461, 2015.

\bibitem{van2009theoretical}
Harm Van~Seijen, Hado Van~Hasselt, Shimon Whiteson, and Marco Wiering.
\newblock A theoretical and empirical analysis of expected sarsa.
\newblock In {\em 2009 ieee symposium on adaptive dynamic programming and
  reinforcement learning}, pages 177--184. IEEE, 2009.

\bibitem{reservoirSampling}
Jeffrey~S Vitter.
\newblock Random sampling with a reservoir.
\newblock {\em ACM Transactions on Mathematical Software (TOMS)}, 11(1):37--57,
  1985.

\bibitem{dueling-dqn}
Ziyu Wang, Nando de~Freitas, and Marc Lanctot.
\newblock Dueling network architectures for deep reinforcement learning.
\newblock {\em CoRR}, abs/1511.06581, 2015.

\bibitem{waugh2015solving}
Kevin Waugh, Dustin Morrill, James~Andrew Bagnell, and Michael Bowling.
\newblock Solving games with functional regret estimation.
\newblock In {\em Twenty-Ninth AAAI Conference on Artificial Intelligence},
  2015.

\bibitem{zinkevich2008regret}
Martin Zinkevich, Michael Johanson, Michael Bowling, and Carmelo Piccione.
\newblock Regret minimization in games with incomplete information.
\newblock In {\em Advances in neural information processing systems}, pages
  1729--1736, 2008.

\end{thebibliography}
